%% file: bmvc_review_final_221003.tex
\documentclass{bmvc2k}

\usepackage{amssymb}
\usepackage{amsfonts}
\usepackage{soul}
\usepackage{lscape}
\usepackage{colortbl}
\usepackage{multirow}
\usepackage{graphicx}
\usepackage{booktabs}
\usepackage{hyperref}
    
\title{
Doubly Contrastive End-to-End Semantic Segmentation for Autonomous Driving under Adverse Weather
}

\addauthor{Jongoh Jeong}{jojeong@rit.kaist.ac.kr}{1}
\addauthor{Jong-Hwan Kim}{johkim@rit.kaist.ac.kr}{1}

\addinstitution{
Korea Advanced Institute of Science and Technology (KAIST)\\
Daejeon, Republic of Korea
}

\runninghead{Jeong, Kim}{Doubly Contrastive End-to-End Semantic Segmentation}

\begin{document}
\maketitle

\input{0.abstract}
\input{1.introduction}

\input{2.relatedwork}
\input{3.method}
\input{4.experiments}

\input{5.conclusion}

\textbf{Acknowledgments.} \quad
This work was supported by the Institute for Information \& Communications Technology Promotion (IITP) under Grant 2020-0-00440 through the Korean Government (Ministry of Science and ICT (MSIT); Development of Artificial Intelligence Technology that Continuously Improves Itself as the Situation Changes in the Real World).

\bibliography{references}
\end{document}

%% file: 0.abstract.tex
\begin{abstract}
    
    Road scene understanding tasks have recently become crucial for self-driving vehicles. In particular, real-time semantic segmentation is indispensable for intelligent self-driving agents to recognize roadside objects in the driving area. As prior research works have primarily sought to improve the segmentation performance with computationally heavy operations, they require far significant hardware resources for both training and deployment, and thus are not suitable for real-time applications. As such, we propose a doubly contrastive approach to improve the performance of a more practical lightweight model for self-driving, specifically under adverse weather conditions such as fog, nighttime, rain and snow. Our proposed approach exploits both image- and pixel-level contrasts in an end-to-end supervised learning scheme without requiring a memory bank for global consistency or the pretraining step used in conventional contrastive methods. We validate the effectiveness of our method using SwiftNet on the ACDC dataset, where it achieves up to 1.34\%p improvement in mIoU (ResNet-18 backbone) at 66.7 FPS (2048$\times$1024 resolution) on a single RTX 3080 Mobile GPU at inference. Furthermore, we demonstrate that replacing image-level supervision with self-supervision achieves comparable performance when pre-trained with clear weather images.
    
\end{abstract}

%% file: 1.introduction.tex
\section{Introduction}
\label{sec:introduction}

    Road scene understanding tasks have recently been popular areas of research, gaining traction with the growth of deep learning and interests in intelligent driving agents. In particular, in the field of autonomous driving, achieving high performance in the task of semantic segmentation is key to recognizing roadside objects in the scene ahead~\cite{zhao2018icnet, long2015fully}. A number of existing semantic segmentation models follow the encoder-decoder structure~\cite{long2015fully, valada2017adapnet, badrinarayanan2017segnet, chen2018deeplabv3plus} or a multi-scale pyramidal encoder followed by upsampling~\cite{DBLP:journals/corr/PaszkeCKC16, SwiftNet_orvsic2021efficient, HRNetV2OCR_wang2020deep, 9399071} to categorize each pixel into the defined set of object classes, with 2D RGB images as input.
    
    However, as intelligent driving systems require safe and accurate perception of the surroundings in dynamically changing environments, most heavyweight models that focus primarily on the performance are not suitable for practical real-time deployment. In addition, although various methods target semantic segmentation of the 19 \textit{Cityscapes}~\cite{cordts2016cityscapes} benchmark classes under clear, daytime weather conditions and have been useful in most \textit{normal} road scenarios, their direct application under adverse driving conditions, or ``unusual road or traffic conditions that were not known" as defined by the US Federal Motor Carrier Safety Administration~\cite{code2022fmcsa} (\textit{e.g.}, fog, nighttime, rain, snow), has yet to be explored further~\cite{Sakaridis2021ACDC}. In this light, a number of models and datasets have been proposed to boost the segmentation performance under such adverse conditions~\cite{dai2020foggyzurich,SDV19darkzurich,SDV18cityscapesfoggy,yu2020bdd100k} as there are 1.3 million death toll of road traffic crashes every year, and the risk of accidents in rainy weather, for example, is 70\% higher than in normal conditions~\cite{stat2_andrey, stat1_who}. In this approach, we seek to adapt and extend the segmentation capability of intelligent self-driving agents to such adverse conditions by exploiting a lightweight model and minimizing additional costs for ground truth labels. 
    
    To this end, we propose a doubly contrastive end-to-end learning method for a lightweight semantic segmentation model under adverse weather scenarios. Given an RGB image as input, our proposed strategy optimizes according to the supervised contrastive objective in both image- and pixel-levels in an end-to-end manner, in addition to the segmentation loss. In contrast to the previous contrastive learning approaches, we target an end-to-end training with direct feature learning, eliminating the pre-training stage in the common two-stage training scheme, and also examine how much of a performance boost image-level labels can further contribute in the supervised semantic segmentation task.
    
    We highlight our main contributions in three-fold:
    \vspace{-0.5em}
    \begin{itemize}
    \itemsep -4pt
        \item We propose an end-to-end doubly (image- and pixel-levels) contrastive learning strategy for a lightweight semantic segmentation model to eliminate the pre-training stage in the conventional contrastive learning approach without requiring a large training batch size or a memory bank.
        \item Our training method achieves 1.34\%p increase in mIoU measure from the baseline focal loss-only objective with the SwiftNet architecture (ResNet-18 backbone), running inference at up to 66.7 FPS in 2048$\times$1024 resolution on a single Nvidia RTX 3080 Mobile GPU.
        \item We verify that replacing image-level supervision with self-supervision in our supervised contrastive objective achieves comparable performance when pre-trained with clear weather images.
    \end{itemize}

%% file: 2.relatedwork.tex
    \section{Related Work}
\label{sec:relatedwork}
 
\textbf{Supervised semantic segmentation.} \quad
    In categorizing each pixel into its representative semantic class, many previous studies have employed the encoder-decoder structure~\cite{badrinarayanan2017segnet,long2015fully,noh2015learning,ronneberger2015unet}. An encoder-decoder network consists of an encoder module that gradually reduces the feature maps to capture enriched semantic features, and a decoder that also gradually proceeds to recover the spatial information. This structure allows for faster computation as it does not make use of dilated features in the encoder and can recover sharp object boundaries in the decoder~\cite{badrinarayanan2017segnet, ronneberger2015unet}. Extending this structure, DeepLabV3+~\cite{chen2018deeplabv3plus} adds a multi-scale aspect in the encoder as well as depth-wise separable convolution and atrous spatial pyramid pooling (ASPP) to obtain sharper segmentation outputs. While it lowers the computational cost overhead, it is still heavy in size (\textit{e.g.,} in floating point operations (FLOPs)) for practical deployment. On the other hand, SwiftNet~\cite{SwiftNet_orvsic2021efficient} uses a real-time, efficient encoder that sequentially extracts and concatenates multi-scale pyramidal features, and ENet~\cite{DBLP:journals/corr/PaszkeCKC16} is an even lighter model which extracts features in a single scale only.

    \noindent \textbf{Representation learning without supervision.} \quad
        %
        %
        The original idea of matching representations of a single image in agreement dates back to \cite{becker1992selfsupervised} and in the direction of preserving consistency in representations, \cite{berthelot2019mixmatch, xie2020unsupervised} advance the idea in semi-supervised settings. Previous attempts to adjust data representations appropriately to a specific task include learning with pretext tasks such as relative patch prediction~\cite{doersch2017multi}, jigsaw puzzle solving~\cite{noroozi2016unsupervised}, colorization~\cite{he2016deep} and rotation prediction~\cite{gidaris2018unsupervised}. 
        %
        A representative approach to learning generalized data representation by contrastive learning is SimCLR by Chen \textit{et al.}~\cite{chen2020simple}, an augmentation-based contrastive learning method for consistent representations. It operates by augmenting the input with two random transformations and maximizing their mutual agreement, thereby achieving comparable performance to the supervised setting in image classification after pre-training on relevant pretext tasks. \cite{chen2020big} extends SimCLR by further enhancing the representation learning capability with a larger model architecture, and validates its effectiveness when fine-tuned or distilled onto another smaller network. 
        
    \noindent \textbf{Representation learning with supervision.} \quad
        SupContrast~\cite{khosla2020supervised} extends SimCLR by directly incorporating image-level labels for the image classification task. 
        In this supervised scheme, pairwise equivalance matching of representations based on their category labels enhances the push-pull effect in the feature space during training. 
        %
        %
        In semantic segmentation, \cite{wang2021exploring} introduces cross-image pixel-wise contrast to explicitly address intra-class compactness and inter-class dispersion phenomena to consider pixel-wise semantic correlations globally.
        However, as most contrastive representation learning methods are limited by a smaller mini-batch size than the number of necessary negative samples in practice, \cite{wang2021exploring} stores all region (pooled) embeddings from training data in an external memory bank, transposing the loss from \textit{pixel-to-pixel} to \textit{pixel-to-region}. Another work in a semi-supervised setting~\cite{alonso2021semi} also uses a memory bank to store class-wise features learned from a teacher network, from which a subset of selected features are used as pseudo-labels for the student network in its pixel-wise contrast. Although both are trainable end-to-end owing to a memory bank, they do require considerable memory resources for storing feature embeddings.
        
    \noindent \textbf{Mixed supervision.} \quad
        Many works have sought to reduce human costs for acquiring labels by leveraging weak labels as well. \cite{xu2015learning} proposes a unified algorithm that learns from various forms of weak supervision including image-level tags, bounding boxes and partial labels to produce pixel-wise labels, 
        and \cite{ahn2018learning, pinheiro2015image} only utilize image-level labels as weak priors to generate accurate segmentation labels. As such, weak labels such as image-level labels have become useful in learning precise mappings from RGB to segmentation maps. 
        
        

%% file: 3.method.tex
\section{Methodology}
\label{sec:methodology}

        
\subsection{Preliminaries}
    
    \textbf{Self-Supervised Contrast.}
    For a set of $N$ randomly sampled image-label pairs, $\{\mathbf{x}_{k}, \mathbf{y}_{k}\}_{k=1\hdots N}$, we prepare a corresponding \textit{multi-viewed} set of augmented samples originating from the same sources, $\{\tilde{\mathbf{x}}_{k},\tilde{\mathbf{y}}_{k}\}_{k=1\hdots 2N}$, where each consecutive $k^{th}$- and $(k+1)^{th}$-index sample pair originates from the same source. 
    We take a batch of data containing two sets of $N$ arbitrarily augmented samples as input to the self-supervised contrastive loss, following~\cite{chen2020simple, khosla2020supervised} which stem from InfoNCE~\cite{gutmann2010noise,van2018representation}. We let $i\in I \equiv \{1,\hdots,2N\}$ be the anchor index, and the corresponding other augmented sample as $j(i)$ (\textit{positive}). For $\mathbf{z}_{i} = Proj(Enc(\tilde{\mathbf{x}}_{i}))\in \mathbb{R}^{D_{proj}}$, we apply the inner (dot) product and softmax operations on the $i$-th and the corresponding $j(i)$-th projected encoded representations, as follows: 
    
    
    \begin{equation}
        \mathcal{L}_{self} = \sum_{i\in I}\mathcal{L}^{(i)}_{self} = -\sum\limits_{i\in I}\log \frac{\exp(\mathbf{z}_{i}\cdot\mathbf{z}_{j(i)}/\tau)}{\sum\limits_{a\in A(i)} \exp(\mathbf{z}_{i}\cdot \mathbf{z}_{a}/\tau)}
        \label{eq:contr_self_supervised}
    \end{equation}
    \noindent where $A(i)\equiv I\setminus {i}$ and the remaining set of images $\{k\in A(i) \setminus j(i)\}$ contains $2(N-1)$ number of \textit{negative} samples. $\tau\in \mathbb{R}^{+}$ is the temperature parameter for the softmax, whereby the higher the temperature value, the softer (lower) the logits output of the softmax function. For each anchor $i$, we use 1 positive pair and $2N-2$ negative pairs, thus in total of $2N-1$ terms to obtain the loss value.

    \noindent \textbf{Supervised Contrast.} \quad
    For \textit{image-level} supervised contrast, there are present more than one sample belonging to each image class label.  We thus take the average of all values over all \textit{positives} for the anchor sample $i$, such that $P(i) \equiv \{p\in A(i) | \tilde{\mathbf{y}}_{p} = \tilde{\mathbf{y}}_{i}\}$. Note that the key difference between self-supervised and supervised contrastive approaches is that for each anchor, multiple \textit{positives} and \textit{negatives} from the same and different classes, respectively, are considered rather than different data augmentations of the same anchor.
    Following the loss objective in \cite{khosla2020supervised} which takes the summation outside the log operator, we define the image-level supervised contrastive loss as follows: 
    
    \begin{equation}
        \mathcal{L}_{image} = \sum_{i\in I}\mathcal{L}^{(i)}_{image} = -\sum\limits_{i\in I} \frac{1}{|P(i)|} \sum\limits_{p\in P(i)} \log \frac{\exp(\mathbf{z}_{i}\cdot\mathbf{z}_{p}/\tau)}{\sum\limits_{a\in A(i)} \exp(\mathbf{z}_{i}\cdot \mathbf{z}_{a}/\tau)}
        \label{eq:contr_supervised_image}
    \end{equation}
    \noindent where $|P(i)|$ denotes the cardinality of the set of \textit{positives}. Note that easy positives and negatives (\textit{i.e.}, the dot product $\sim$ 1) contribute to the gradient relatively smaller than the hard positives and negatives (\textit{i.e.}, the dot product $\sim$ 0). We refer to \cite{khosla2020supervised} for detailed proofs. 


    \textit{Pixel-level} supervised contrastive loss follows a similar form as the image-level, except that the representation is on the each pixel rather than the entire sample, and \textit{positives} and \textit{negatives} come from the same image rather than from two randomly augmented images. As in Wang \textit{et al.}~\cite{wang2021exploring}, the pixel-wise contrast loss addresses two limitations in using the cross-entropy (CE) loss, in which (1) it penalizes predictions neglecting the pixel-wise semantic correlations, and (2) the relative relations among pixel-wise logits fail to directly supervise on the features. 
    %
    We formulate the pixel-level supervised contrastive loss as follows:
    %
    \begin{equation}
        %
        \mathcal{L}_{pixel} = -\frac{1}{|P(i)|} \sum\limits_{i_{p}\in P(i)} \log \frac{\exp(\mathbf{i}\cdot\mathbf{i}_{p}/\tau)}{\sum\limits_{i_{a}\in A(i)} \exp(\mathbf{i}\cdot\mathbf{i}_{a}/\tau)}
        \quad\quad\quad
        \forall i\in\{1,\hdots,H\times W\},
        \label{eq:contr_supervised_pixel}
    \end{equation}
    \noindent where $A(i)\equiv Q\setminus i$, in which $Q$ represents the set of all pixels in the multi-viewed data batch and $i$ denotes the pixel anchor index in a given image. $P(i)$ denotes pixel embedding collections of the \textit{positive} samples for each pixel $i$ in an image of size $H\times W$.

\subsection{Doubly Contrastive Supervised Segmentation}

    \begin{figure}[!h]
        \centering
        \includegraphics[width=0.9\linewidth]{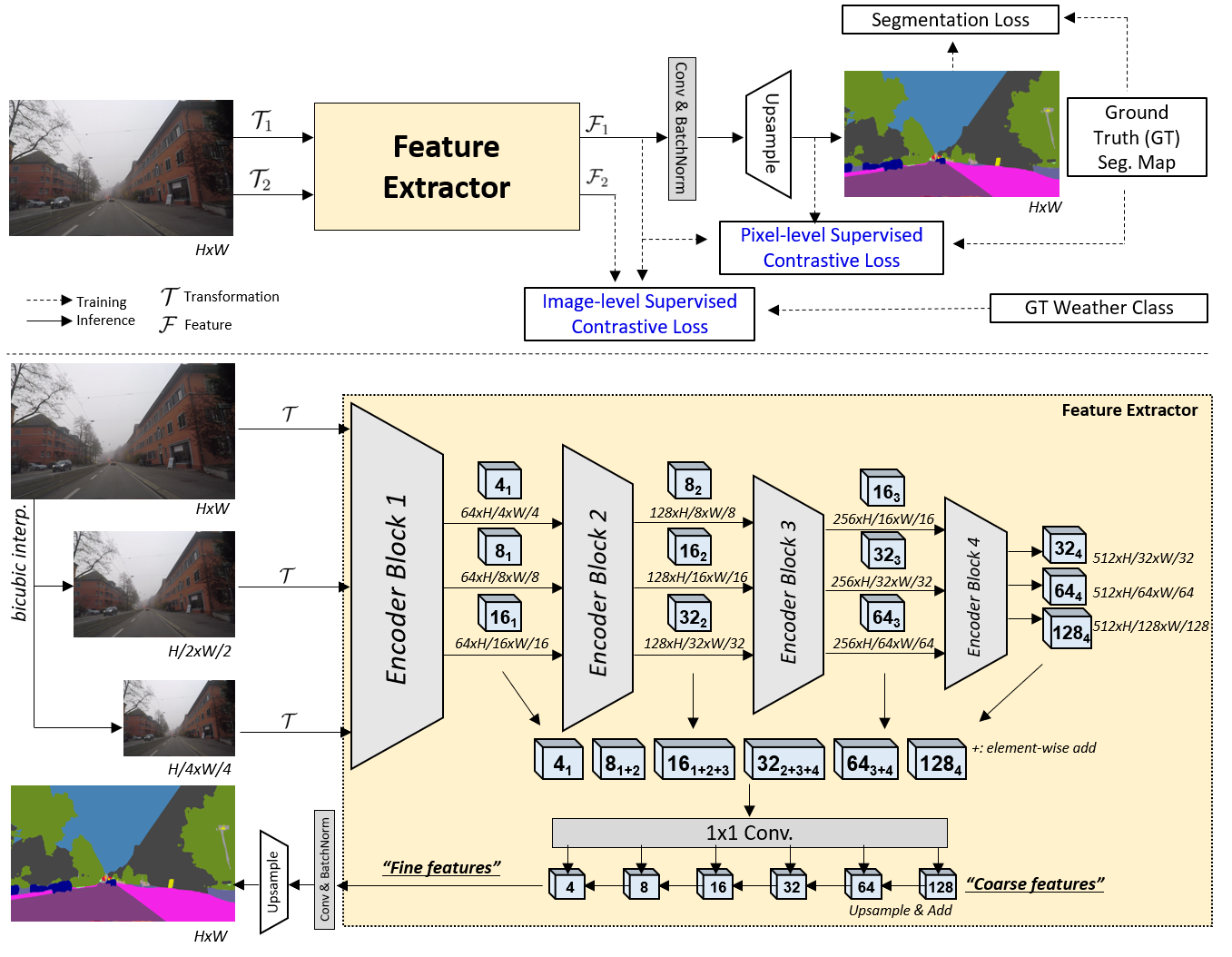}
        \caption{Overview of the training pipeline (top) and SwiftNet architecture (bottom).}
        \label{fig:training_protocol}
    \end{figure}


    

    


    We remark that our approach as shown in Fig.~\ref{fig:training_protocol} is, in fact, also applicable to other supervised semantic segmentation tasks in which we have available ground truth pixel-level semantic labels as well as the corresponding image-level labels for those pixels in each image. Adverse weather conditions is one such case where we can readily acquire and exploit image-level contexts to promote contextual contrastive learning under unfavorable outdoor weather conditions.

    
    The base model for our experiments is SwiftNet~\cite{SwiftNet_orvsic2021efficient} with the ResNet-18 backbone to balance the trade-off between the accuracy and the scale of model parameters and computations. SwiftNet is advantageous for its light weight and effectiveness in self-driving scenarios that often require real-time inference speed as well as comparably high performance to those of large-scale models~\cite{song2022rodsnet}. Moreover, its multi-scale pyramidal features extracted from the sequential encoder blocks provide scale-invariant features that capture spatial information more precisely than single-scale models~\cite{kim2018fpn}.

        \textbf{Loss objective.} \quad
            Standard semantic segmentation networks have long been trained according to the pixel-wise CE loss given class-specific probability distributions by weighting all pixels equally and considering the frequency of pixels belonging to each class~\cite{SwiftNet_orvsic2021efficient,zhang2019deep}. This approach, however, is prone to overfitting and biased towards more frequent classes. In addition, the CE loss suffers from failing to recognize edges in pixel-level granularity. To remedy these problems, \cite{lin2017focal, SwiftNet_orvsic2021efficient} have employed class balancing by adopting a scaling coefficient term to weight each class in a less biased and balanced manner. 
            
            We thus follow \cite{song2022rodsnet} in order to weight the cross-entropy and the focal loss~\cite{lin2017focal} with the class-balancing term as described in Eq.~\ref{eq:seg_loss}, where we take the inverse logarithm of the ratio of the frequency of each pixel appearance, $freq_{c}$, over the entire pixels in the dataset, $N_{p}$. We choose $\epsilon=1\times 10^{-1}$ for numerical stability.
            The Euclidean distance transform (EDT)~\cite{felzenszwalb2012EDT} serves to weight each pixel by the distances to the boundaries of other objects. That is, for each pixel $p$ in a given image grid $G_{H\times W}$, we compute the $L_{2}$ distance to the nearest unmasked pixel $q$, \textit{i.e.,} the closest valid pixel belonging to the set of classes, $C$ as in Eq.~\ref{eq:seg_loss_edt}. With both EDT and class-balancing terms in effect, the pixels of smaller objects and object boundaries are weighted greater than more frequently seen pixels such as those of \textit{sky} or \textit{road} classes during training. 
            
            \begin{align}
                \mathcal{L}_{seg}\left(\phi(p), \hat{\phi}(p)\right) = -\delta(p) e^{\gamma(1-P_{t})}log(P_{t}),
                \label{eq:seg_loss}
            \end{align} \\
            \noindent with
            \[
                \delta(p) =  \underbrace{log\left(1+\epsilon+\frac{freq_{c}(p)}{N_{p}}\right)^{-1}}_{\text{Class balancing}} \cdot
                \underbrace{exp\left(-\frac{d_{EDT}(p)}{2\sigma_{EDT}}\right)}_{\text{EDT}},
                \tag{\theequation .a}\label{eq:seg_loss_delta}
            \]
            \[
                \hspace{-0.83in}
                d_{EDT} (p) = \sum_{C} \min_{q \in C} \left\lVert p-q \right\rVert_{2} \quad\quad\quad \forall p\in G_{H\times W}, \\
                \tag{\theequation .b}\label{eq:seg_loss_edt}
            \]
            \noindent where $P_{t}$ denotes the softmax probability for each semantic class, and $\phi(p)$ and $\hat{\phi}(p)$ are the ground truth and the predicted segmentation maps, respectively. 
            We then combine the segmentation and the contrastive loss terms as the final loss objective as follows:
            \begin{equation}
                \mathcal{L} = \lambda_{c}\cdot (\mathcal{L}_{\text{image}} + \mathcal{L}_{\text{pixel}}) + \lambda_{s}\cdot \mathcal{L}_{seg},
            \end{equation}
            \noindent where $\lambda_{c} = 1/\mathcal{B}$, $\lambda_{s} = 1.2$ and $\mathcal{B}$ is the batch size. 
            For $\mathcal{L}_{image}$ and $\mathcal{L}_{pixel}$, we further add stability during training by adding $L_{2}$ normalization to the inner product (\textit{i.e.,} $\lVert\mathbf{z}_{i}\cdot\mathbf{z}_{p}\rVert_{2}$) to directly mimic the cosine similarity. The scaling coefficients, $\lambda_{c}$ and $\lambda_{s}$, are experimentally determined so as to reduce the unfavorable effect of feature equivalence matching and place a greater weight upon the segmentation objective. We also scale the contrastive loss value empirically to be close to the segmentation loss value. In this combined objective, the image-level contrast complements the pixel-level contrast such that the semantic correlations among intra-image pixels of each image are consistent across images of different weather classes globally without a memory bank. For the following experiments that use self-supervision in place of image-level contrast, we simply replace $\mathcal{L}_{image}$ with $\mathcal{L}_{self}$.

%% file: 4.experiments.tex
\section{Experiments}
\label{sec:experiments}

\subsection{Datasets and Evaluation Metric}
    
    We evaluate the proposed method using the training and validation sets of an urban street scene dataset, the Adverse Conditions Dataset with Correspondences for Semantic Driving Scene Understanding (ACDC)~\cite{Sakaridis2021ACDC}, with and without pre-training on another driving scene benchmark dataset, the Cityscapes~\cite{cordts2016cityscapes} as described in Table~\ref{tab:datasets}. 
    
    Cityscapes is a road scene benchmark dataset containing stereo RGB images taken from the egocentric point of view of the driver under clear, daytime weather, with pixel-level annotations for 19 semantic classes of various roadside objects including bus and fence. 
    ACDC is a smaller dataset labelled according to the Cityscapes categories, yet in adverse conditions (fog, nighttime, rain and snow). While other Cityscapes-like datasets~\cite{SDV18cityscapesfoggy, dai2020foggyzurich, SDV19darkzurich, zendel2018wilddash, dai2018nightdriving, yu2020bdd100k, tung2017raincouver} contain few to several hundred images biased toward specific adverse conditions, \textit{ACDC} offers evenly-balanced number of images and the corresponding labels.
    
    We evaluate the semantic segmentation performance by the mean Intersection-over-Union (IoU), also known as the Jaccard Index.
    IoU is often computed using the confusion matrix, from which we obtain true positive (TP), false positive (FP) and false negative (FN) scores for the predictions on the validation set, \textit{i.e.,  IoU = TP / (TP+FP+FN)}. The average IoU score computed over all fine category labels is denoted by mIoU, neglecting the \textit{void} label. 

    \begin{table}[!h]
        \centering
        \caption{Dataset Description}
        \label{tab:datasets}
        \resizebox{\textwidth}{!}{
            \begin{tabular}{cccccccc}
                \toprule
                \textbf{Dataset} & \textbf{Input Modality} & \textbf{Resolution} & \textbf{Anno. (\# Classes)} & \textbf{\shortstack{Weather\\Condition}} & \textbf{Train} & \textbf{Val} & \textbf{Test} \\
                \midrule
                
                Cityscapes~\cite{cordts2016cityscapes} & Stereo RGB & 2048$\times$1024 & Fine (19) & Clear/Daytime & 2,975 & 500 & 1,525 \\
                \midrule
                
                \multirow{5}{*}{ACDC~\cite{Sakaridis2021ACDC}} & \multirow{5}{*}{Monocular RGB} & \multirow{5}{*}{1920$\times$1080} & \multirow{5}{*}{Fine (19)} & All & 1,600 & 406 & 2,000 \\ \cmidrule{5-8}
                & & & & Fog & 400 & 100 & 500 \\
                & & & & Nighttime & 400 & 106 & 500 \\
                & & & & Rain & 400 & 100 & 500 \\
                & & & & Snow & 400 & 100 & 500 \\
                
                \bottomrule
            \end{tabular}
        }
    \end{table}
        
    


\subsection{Implementation Details}
    %
    %
    We performed our experiments on a single Nvidia RTX 3090 GPU with PyTorch 1.7.1, CUDA 11.1, and CUDNN 8.3.2. In order to account for scale variations, we augmented the data with the following transformation: random square crop (768$\times$768) followed by scaling by a random value sampled from a uniform distribution$\sim$$U(0.5,2.0)$. We did not apply transformations that could potentially harm the image quality in adverse weather conditions such as color jitter. We set the validation image width and height to those of \textit{Cityscapes} (2048$\times$1024) for fair comparison of the segmentation results. We pre-trained each model with ImageNet~\cite{ILSVRC15} and set $\gamma=0.5$ in the focal loss, the temperature $\tau=0.07$, and the feature dimension to $D_{proj}=128$ for the supervised contrast. We trained the network on the ACDC dataset for 400 epochs with a batch size of 8, using an Adam~\cite{kingma2014adam} optimizer with $\beta_{1}=0.9$, $\beta_{2}=0.99$. We set the initial learning rate and the weight decay to $4\times10^{-4}$ and $1\times10^{-4}$, respectively, and used a cosine annealing scheduler to decay the learning rate to $1\times10^{-6}$ in the last epoch.

\subsection{Evaluation}
    \noindent \textbf{Quantitative results.} \quad
    We summarize the semantic segmentation results in Table~\ref{tab:semseg_predictions_swiftnet}, where there are 1.34\%p and 1.33\%p increases in mIoU from the baseline for the ResNet-18 and -34 backbones, respectively, for our method. Our proposed method generally outperforms in mIoU under each of the four adverse conditions, relative to the other experimented single and double contrasts. Our method is particularly effective for the \textit{harder} conditions, namely \textit{nighttime}, \textit{rain} and \textit{snow}, where we exploit the semantic correlations in the embeddings across the pixels of different conditions owing to the doubly contrastive objective. We remark that the difficulties in predictions in nighttime and rain stem from a significant number of darker pixels obstructing clear view of objects in the driving area and rain droplets distorting the camera view and the texture of objects at far distances, respectively. 
    The performance under \textit{fog} is not relatively lower than in clear weather due to low fog density.
    Further, we highlight that our experiments are accompanied with a batch size of 8 under a more practical training scenario. We expect higher performance gain with a batch size of 2048 as suggested in common contrastive learning works~\cite{khosla2020supervised}. 
    
    With pre-training on clear weather images, there are 0.86\%p and 1.80\%p improvement in mIoU for the ResNet-18 and -34 backbones for our method, respectively, after fine-tuning on the adverse weather images. Note that the experiment (f) achieves as high overall mIoU as the experiment (g), as well as in each weather-specific mIoU score. While our method has minor improvements for all conditions except fog with the ResNet-18, we observed noticeable increases with the ResNet-18 backbones.

\begin{figure}[!t]
        \centering
        \includegraphics[width=\linewidth]{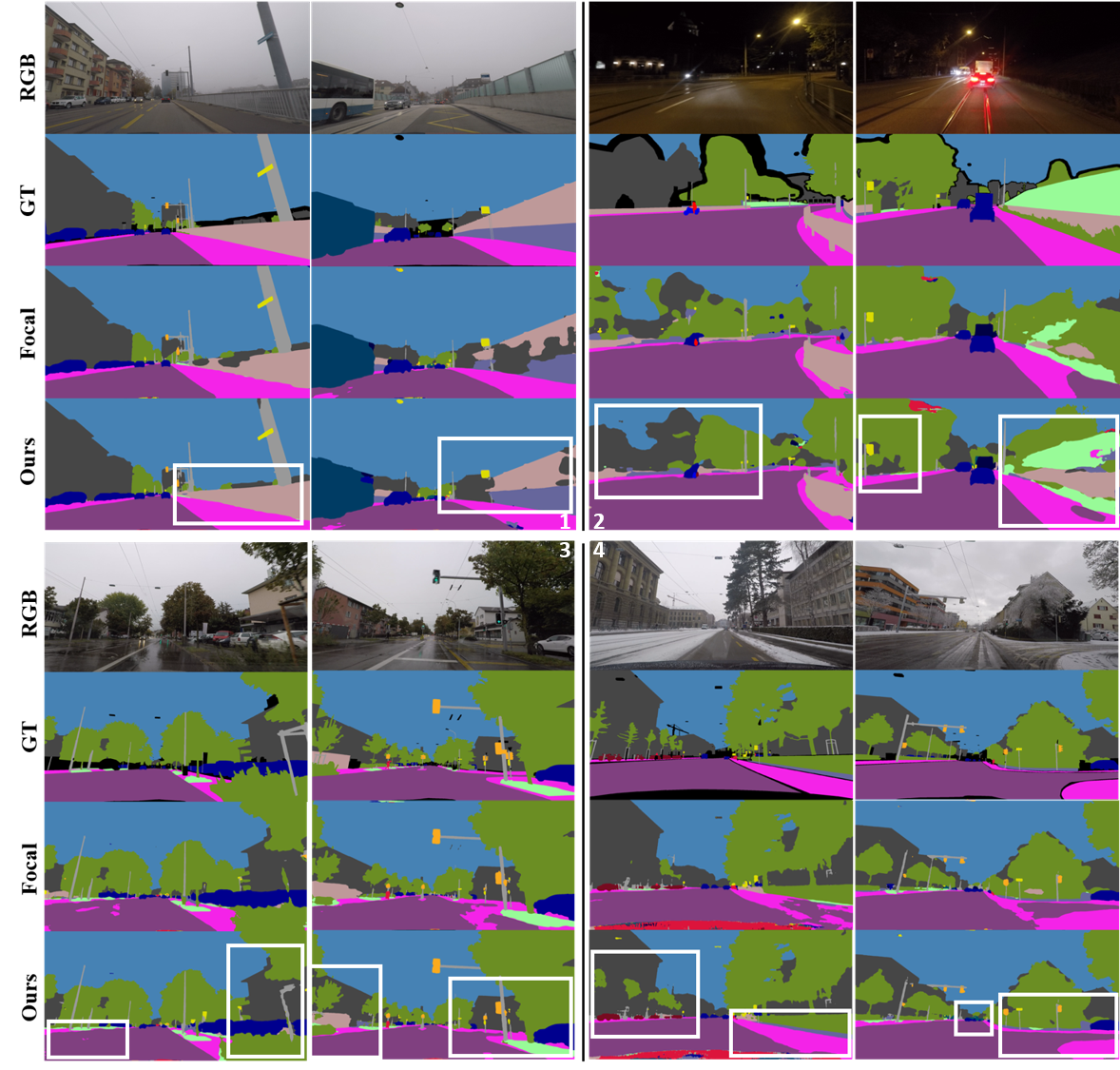}
        \caption{Two samples of semantic segmentation prediction results for each of the four weather conditions: fog, night, rain and snow (1 $\rightarrow$ 4). \textit{Please zoom in to see details.}}
        \label{fig:semseg_predictions}
    \end{figure}

    \noindent \textbf{Qualitative results.} \quad
    As shown in Fig.~\ref{fig:semseg_predictions}, our doubly contrastive method corrects false positive predictions from the focal loss (baseline) case, and allows to predict objects of a relatively large size more consistently. For instance, it removes noisy predictions on \textit{road}, removes the fuzzy boundary between \textit{road} and \textit{sidewalk}, and fills in mis-classified pixels with the correct label for somewhat noisily predicted objects.
    Specifically in \textit{rain}, it recovers a \textit{pole} that seemed nonexistent in the baseline and corrects the pixels mis-classified as \textit{sidewalk} to \textit{road}. The effect of our method is more apparent when there is a region with a majority of pixels belonging to a single class, yet there remains mis-classified or noisy pixels in part.
    
    \begin{table}[!t]
        \centering
        \caption{Semantic segmentation performance by adverse weather conditions using SwiftNet. Bb denotes backbone. The best results in \textbf{boldface} and the second best in \underline{underline}.}
        \label{tab:semseg_predictions_swiftnet}
        \resizebox{0.95\textwidth}{!}{
            \begin{tabular}{cclc|ccccc}
                \toprule
                \textbf{Bb} & \textbf{Exp.} & \textbf{Loss} \footnotesize{($^{*}$: single contrast, $^\dag$: double contrasts)} & \textbf{mIoU} (\%) & \textbf{Fog} & \textbf{Nighttime} & \textbf{Rain} & \textbf{Snow} \\
                \midrule
                \multirow{7}{*}{\rotatebox[origin=c]{90}{ResNet-18}} & (a) & Cross Entropy & 62.63 & 68.95 & 45.71 & 63.32 & 65.43 \\ 
                & (b) & Focal (baseline) & 64.04             & 71.59             & 47.84             & 63.90 & 65.57 \\ 
                & (c) & +Pixel-level Supervised Contrast only$^{*}$ & 63.79             & 69.76             & 47.26             & 63.05             & 68.34 \\ 
                & (d) & +Self-supervised Contrast only$^{*}$ & 63.91             & \underline{71.83} & 48.02             & 62.24             & \underline{68.35} \\ 
                & (e) & +Image-level Supervised Contrast only$^{*}$ & 63.62             & 69.66             & 47.65             & 63.28             & 67.10 \\ 
                & (f) & +Self-supervised and Pixel-level Supervised Contrasts$^\dag$ & \underline{65.07} & \textbf{72.45}    & \textbf{48.57}    & \underline{63.95}             & 68.31 \\ 
                & (g) & +Image- and Pixel-level Supervised  Contrasts$^\dag$ \textit{(Ours)} & \textbf{65.38}    & 67.94             & \underline{48.56} & \textbf{65.38}    & \textbf{68.64} \\ 
                \midrule
                
                \multirow{7}{*}{\rotatebox[origin=c]{90}{ResNet-34}} & (a) & Cross Entropy & 65.02 & 73.85 & 47.11 & 65.20 & 66.94 \\ 
                & (b) & Focal (baseline) & 67.00             & 74.44             & 49.38             & \underline{67.29} & 70.64 \\ 
                & (c) & +Pixel-level Supervised Contrast only$^{*}$ & 66.60             & 72.49             & 49.84             & 65.55             & 69.06 \\ 
                & (d) & +Self-supervised Contrast only$^{*}$ & \underline{68.09} & \textbf{76.99}    & 50.14             & 66.45             & 69.56  \\ 
                & (e) & +Image-level Supervised Contrast only$^{*}$ & 68.07             & 75.12 & \underline{50.76} & 66.78             & 69.33 \\ 
                & (f) & +Self-supervised and Pixel-level Supervised Contrasts$^\dag$ & 67.02             & 72.06             & 50.12             & 65.86             & \underline{70.83} \\ 
                & (g) & +Image- and Pixel-level Supervised Contrasts$^\dag$ \textit{(Ours)} & \textbf{68.33}    & \underline{75.19}             & \textbf{51.21}    & \textbf{67.37}    & \textbf{71.19} \\ 
                
                \midrule
                \midrule
                
                \textbf{Bb} & \textbf{Exp.} & \textbf{Loss} (w/ pre-training with clear weather) & \textbf{mIoU} (\%) & \textbf{Fog} & \textbf{Nighttime} & \textbf{Rain} & \textbf{Snow} \\
                \midrule
                
                \multirow{5}{*}{\rotatebox[origin=c]{90}{ResNet-18}} & - &  \multicolumn{1}{l}{Focal (\textit{Cityscapes})} & 73.16 & \multicolumn{4}{c}{N/A (Clear weather only)} \\ \cmidrule{2-8}
                & (a) & Cross Entropy & 64.40 & 71.88 & 47.09 & 65.59 & 66.66 \\ 
                & (b) & Focal (baseline) & 65.49 & 73.43 & 47.67 & 64.77 & 68.35 \\ 
                & (f) & +Self-supervised and Pixel-level Supervised Contrasts$^\dag$ & \textbf{66.97} & \underline{74.05} & \textbf{49.10} & \textbf{67.85} & \textbf{69.69} \\ 
                & (g) & +Image- and Pixel-level Supervised Contrasts \textit{(Ours)}$^\dag$ & \underline{66.24} & \textbf{75.38} & \underline{48.82} & \underline{65.79} & \underline{68.42} \\ 
                \midrule
                
                \multirow{5}{*}{\rotatebox[origin=c]{90}{ResNet-34}} & - & \multicolumn{1}{l}{Focal (\textit{Cityscapes})} & 73.80 & \multicolumn{4}{c}{N/A (Clear weather only)} \\ \cmidrule{2-8}
                & (a) & Cross Entropy & 68.38 & 75.99 & 49.70 & 67.92 & 71.49 \\
                & (b) & Focal (baseline) & 69.46 & \textbf{76.94} & 50.28 & 69.78 & 71.69 \\ 
                & (f) & +Self-supervised and Pixel-level Supervised Contrasts$^\dag$ & \underline{70.06} & 76.11 & \underline{50.47} & \textbf{70.68} & \textbf{72.89} \\ 
                & (g) & +Image- and Pixel-level Supervised Contrasts \textit{(Ours)}$^\dag$ & \textbf{70.13} & \underline{76.31} & \textbf{53.59} & \underline{70.50} & \underline{71.89} \\ 
                
                \bottomrule
            \end{tabular}
        }
    \end{table}

    \noindent \textbf{Feature visualization.} \quad
    \begin{figure}[!t]
        \centering
        \includegraphics[width=\textwidth]{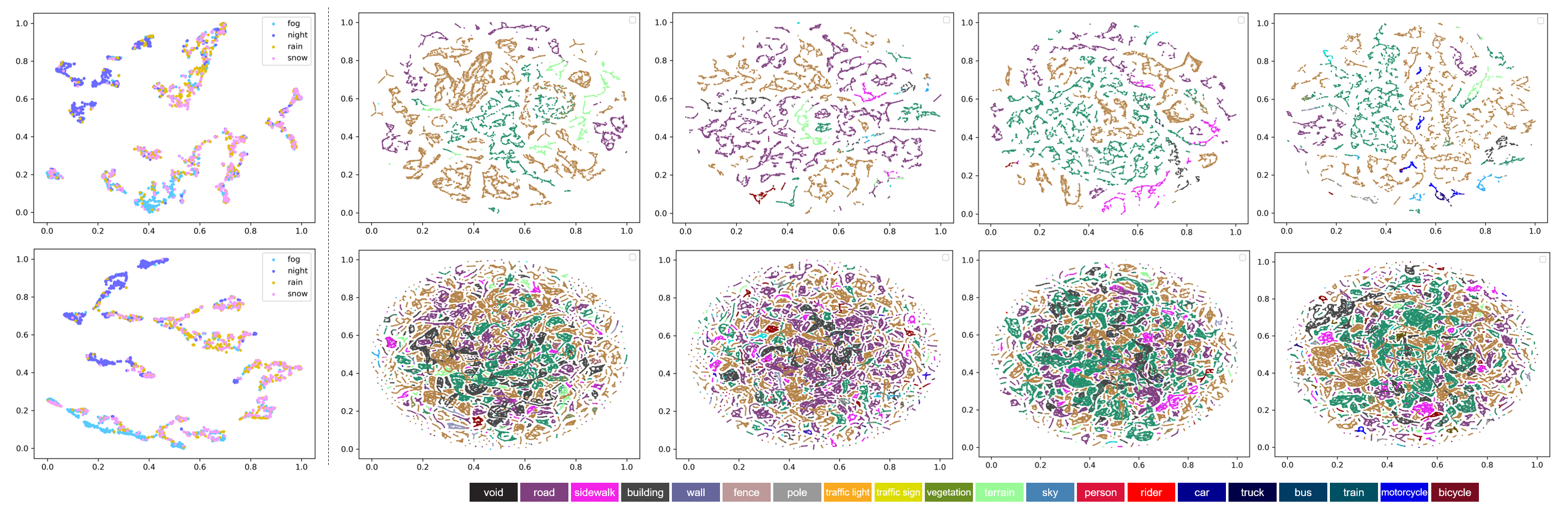}
        \caption{t-SNE visualizations for the trained features using SwiftNet (ResNet-18). Top row: focal loss only (baseline). Bottom row: Ours. Columns from left to right: weather class-wise features, and pixel-wise features for fog, nighttime, rain and snow, in order. \textit{Void} class is not shown. \textit{Please zoom in to see details.}}
        \label{fig:feature_viz}
    \end{figure}
    For an in-depth analysis of the learned features, we provide the class- and pixel-wise t-SNE visualizations for the baseline and our method in Fig.~\ref{fig:feature_viz}. For the class-wise features categorized into the four weather conditions, our doubly contrastive method better discriminates images of the \textit{fog} and \textit{night} classes from images of the other two. Images of these two categories are more clustered and densely located by themselves. 
    
    For pixel-wise features, we opted to visualize for the first batch of pixel-wise feature embeddings due to a memory contraint, where we observed that the features are more densely populated, thereby appearing darker in each respective color. This insinuates pixels belonging to the same class are more compactly located in the feature space. While grouping patterns seem more apparent for the baseline, our combination of image- and pixel-level supervised contrastive objectives finds a good balance in the feature space that yields higher semantic segmentation scores.

\noindent \textbf{Ablation studies.} \quad
    We examine the effect of different feature extractors as well as the \textit{fine} vs. \textit{coarse} feature granularity as noted in Fig.~\ref{fig:training_protocol}. Table~\ref{tab:ablation_study} presents the results and the corresponding computational costs when the SwiftNet feature extractor is replaced by that of ENet and DeepLabV3+ (ResNet-50 backbone). While our method is effective with SwiftNet, it is not as effective as with ENet or with DeepLabV3+.
    We infer that this is due to the single-scale encoder for ENet, and the ASPP module for DeepLabV3+. While SwiftNet draws multi-resolution features sequentially from the previous scale features, DeepLabV3+ applies convolutions with different kernel sizes and pools altogether in one step. We applied our image-level contrast to the ENet features before upsampling and DeepLabV3+ features after ASPP followed by 1$\times$1 convolution, respectively. In terms of memory, ENet is the smallest in size, yet it requires heavy memory space and is thus relatively slower. DeepLabV3+, on the other hand, is large in size even with ResNet-50 instead of its proposed Xception-71. Replacing the \textit{fine} features with the \textit{coarse} in SwiftNet yields a drop of 3.75\%p in mIoU for our method since \textit{fine} features contain the fused output from the features from multi-resolution encoder blocks that provide geometrically richer contexts than the \textit{coarse}.
    
    \begin{table}[!th]
        \centering
        \caption{Ablation study: semantic segmentation performance with different models and \textit{coarse} features (2048$\times$1024 resolution). \textit{Coarse} features are marked with $\dag$; otherwise \textit{fine}.}
        \label{tab:ablation_study}
        \resizebox{\textwidth}{!}{
            \begin{tabular}{ccc|cccc|cccc}
                \toprule
                \multirow{2}{*}{\textbf{Model} (Encoder type)} & \multirow{2}{*}{\textbf{Exp.}} & \multirow{2}{*}{\textbf{mIoU} (\%)} & \multirow{2}{*}{\textbf{Fog}} & \multirow{2}{*}{\textbf{Nighttime}} & \multirow{2}{*}{\textbf{Rain}} & \multirow{2}{*}{\textbf{Snow}} & \multirow{2}{*}{\textbf{GFLOPs}} & \multirow{2}{*}{\textbf{Params} (M)} & \multicolumn{2}{c}{RTX 3080 Mobile}\\
                & & & & & & & & & \textbf{Time} (ms) & \textbf{FPS} \\
                \midrule
                
                 & (a) & 45.22 & 48.10 & 36.31 & 44.46 & 47.53 & \multirow{4}{*}{1.40} & \multirow{4}{*}{0.35} & \multirow{4}{*}{31} & \multirow{4}{*}{32.3} \\
                \textbf{ENet}~\cite{DBLP:journals/corr/PaszkeCKC16} & (b) & \underline{50.45} & \underline{55.14} & \underline{38.70} & 49.44 & \textbf{53.44} & & & & \\ 
                (Single-scale) & (f) & \textbf{50.78} & \textbf{55.91} & \textbf{38.96} & \textbf{50.88} & \underline{52.39} & & & & \\ 
                & (g) & 49.32 & 53.64 & 37.85 & \underline{49.86} & 51.17 & & & & \\ 
                \midrule
                
                 & (a) & 62.63 & 68.95 & 45.71 & 63.32 & 65.43 & \multirow{6}{*}{8.04} & \multirow{6}{*}{12.04} & \multirow{6}{*}{15} & \multirow{6}{*}{66.7}\\
                 & (b) & 64.04 & \underline{71.59} & 47.84 & \underline{63.90} & 65.57 & & & & \\
                \textbf{SwiftNet \footnotesize{(ResNet-18)}}~\cite{SwiftNet_orvsic2021efficient} & (f) & \underline{65.07} & \textbf{72.45} & \textbf{48.57} & 63.95 & \underline{68.31} & & & & \\
                (Multi-scale pyramidal) & (g) & \textbf{65.38} & 67.94 & \underline{48.56} & \textbf{65.38} & \textbf{68.64} & & & & \\ 
                \cmidrule{2-7}
                & (f)$^\dag$ & 60.62 & 65.84 & 45.35 & 61.20 & 64.79 & & & & \\
                & (g)$^\dag$ & 61.66 & 67.86 & 46.07 & 62.10 & 64.84 & & & & \\ 
                \midrule
                
                 & (a) & 65.02 & 73.85 & 47.11 & 65.20 & 66.94 & \multirow{4}{*}{14.40} & \multirow{4}{*}{22.15} & \multirow{4}{*}{26} & \multirow{4}{*}{38.5}\\
                \textbf{SwiftNet \footnotesize{(ResNet-34)}}~\cite{SwiftNet_orvsic2021efficient} & (b) & 67.00 & \underline{74.44} & 49.38 & \underline{67.29} & 70.64 & & & & \\ 
                (Multi-scale pyramidal) & (f) & \underline{67.02} & 72.06 & \underline{50.12} & 65.86 & \underline{70.83} & & & \\
                & (g) & \textbf{68.33} & \textbf{75.19} & \textbf{51.21} & \textbf{67.37} & \textbf{71.19} & & & & \\
                \midrule
                
                 & (a) & \underline{69.69} & 73.60 & \underline{51.45} & \underline{72.18} & \underline{72.29} & \multirow{4}{*}{29.88} & \multirow{4}{*}{39.76} & \multirow{4}{*}{14} & \multirow{4}{*}{71.4}\\
                \textbf{DeepLabV3+ \footnotesize{(ResNet-50)}}~\cite{chen2018deeplabv3plus} & (b) & \textbf{70.07} & \textbf{75.18} & \textbf{52.49} & \textbf{73.22} & 71.50 & & & & \\
                (ASPP) & (f) & 69.22 & 73.95 & 50.54 & 70.42 & \textbf{73.24} & & & & \\
                & (g) & 69.04 & \underline{74.54} & 51.20 & 70.70 & 71.83 & & & & \\
                
                \bottomrule
                
            \end{tabular}
        }
    \end{table}

%% file: 5.conclusion.tex
\section{Conclusion}
\label{sec:conclusion}

We proposed an end-to-end doubly contrastive learning approach to semantic segmentation for self-driving under adverse weather. Our doubly contrastive method exploits image-level labels to semantically correlate RGB images taken under various weather conditions and pixel-level labels to obtain more semantically meaningful representations. In our method, the two supervised contrasts complement each other to effectively improve the performance of a lightweight model, without a need for pre-training or a memory bank to associate images across various weather conditions for global consistency. We hope this sheds further light on contrastive learning approaches for real-time deployment of self-driving systems.


